\documentclass{article}

\usepackage[final]{corl_2019} 

\usepackage{graphics} 
\usepackage{epsfig} 
\usepackage{amsmath} 
\usepackage{amssymb}  
\usepackage{bm}  
\usepackage{subcaption}
\usepackage[font=small,labelfont=small]{caption}
\usepackage{hyperref}
\usepackage{wasysym}
\usepackage{cite}

\usepackage[toc,page]{appendix}

\newcommand\trsp{{\scriptscriptstyle\top}}
\newcommand{\ty}[1]{{\scriptscriptstyle{\mathcal{#1}}}}

\newcommand{\indCl}{\ell}
\newcommand{\indData}{n}

\graphicspath{{Figures/}}

\title{Learning from demonstration with model-based Gaussian process}

\author{
  No\'emie Jaquier$^{1}$ \hspace{0.4cm} David Ginsbourger$^{1,2}$ \hspace{0.4cm} Sylvain Calinon$^{1}$\\
  $^1$Idiap Research Institute \hspace{3.2cm} $^2$IMSV, University of Bern\\
  Rue Marconi 19, 1920 Martigny, Switzerland \hspace{0.2cm} Alpeneggstrasse 22, 3012 Bern, Switzerland\\
  \hspace{0.8cm}\texttt{name.surname@idiap.ch} \hspace{1.9cm} \texttt{david.ginsbourger@stat.unibe.ch} \\
  \vspace{-1cm}
}

\begin{document}
\maketitle

\begin{abstract}
In learning from demonstrations, it is often desirable to adapt the behavior of the robot as a function of the variability retrieved from human demonstrations and the (un)certainty encoded in different parts of the task.
In this paper, we propose a novel multi-output Gaussian process (MOGP) based on Gaussian mixture regression (GMR). The proposed approach encapsulates the variability retrieved from the demonstrations in the covariance of the MOGP. Leveraging the generative nature of GP models, our approach can efficiently modulate trajectories towards new start-, via- or end-points defined by the task. Our framework allows the robot to precisely track via-points while being compliant in regions of high variability. We illustrate the proposed approach in simulated examples and validate it in a real-robot experiment.
\end{abstract}



\section{Introduction}
\label{sec:Intro}
In the context of learning from demonstrations (LfD), robot motions can be generated from demonstrated trajectories using various probabilistic methods, e.g. Gaussian mixture regression (GMR) \citep{Calinon2016}, dynamical movement primitives (DMP)~\citep{Pastor09}, probabilistic movement primitives (ProMP) \citep{Paraschos13} kernelized movement primitives (KMP) \citep{Huang18} or Gaussian process regression (GPR) \citep{Schneider10}.
The covariance matrices of the prediction distributions computed by GMR, ProMPs and KMP encode the variability of the predicted trajectory. This variability, reflecting the dispersion of the data collected during the demonstrations, carries important information for the execution of the task. For example, the phases of the task in which a high precision is required, e.g. picking an object in a specific location, are characterized with a low variability, and vice-versa. During the reproduction, the variability is typically used to define robot tracking precision gains and permits the combination of different controllers \citep{Silverio18}. However, the approaches encoding variability do not take into account the availability of data in the different phases of the task. Inversely, the covariance matrices of the prediction distribution of GPs correspond to the prediction uncertainty, which reflects the presence or absence of training data in different phases of the task. This uncertainty measurement has been used, for example, to modulate the behavior of the robot far from the training data \citep{Silverio18} or to actively make requests for new demonstrations in unexplored regions of the input space \citep{Maeda17}.

In LfD, it is often desirable to precisely refine parts of the demonstrated trajectories (e.g. due to changes in the environment), while maintaining the general trajectory shape (mean and variability) as in the demonstrations. It is also desirable to adapt the behavior of the robot, e.g. its compliance at different phases of the tasks, as a function of the variability of the demonstrations or the presence of (un)certainty in the reproduction.
As none of the aforementioned methods provide both information features simultaneously, several approaches have been developed to take into account prediction uncertainty and variability.
In \citep{Schneider10}, the reproduced trajectories are computed as a product of the predictions of local GPs, obtained by clustering the input space similarly to the approach of \citep{Nguyentuong08}. Therefore, by adapting the parameters of each GP, the resulting uncertainty is adapted as a function of the variability of the different phases of the demonstrations.
In \citep{Umlauft2017}, the prediction uncertainty and variability are inferred separately. The trajectories are predicted using a combination of GP and dynamical movement primitives (DMP), therefore providing uncertainty measurement. On the other hand, the variability in the reproduction is determined by inferring the components of the corresponding covariance matrix with GPs.

In this paper, we propose an approach that aims at encapsulating the variability information of the demonstrations in the prediction uncertainty. We take inspiration from multi-output Gaussian processes (MOGPs) under the linear model of coregionalization (LMC) assumption to design a non-stationary, multi-output kernel based on GMR. In contrast with the aforementioned approaches, both variability and uncertainty information are encoded in a single GP. Moreover, we define the prior mean of the process as equal to the mean provided by GMR. This permits to ignore the training data in the generation of new trajectories and to consider only via-points constraints as observed data, therefore alleviating the computational cost of the GP. Moreover, setting the tracking precision as a function of the retrieved covariance allows us to demand the robot to precisely track the via-points while lowering the required tracking precision in regions of high variability. 

The remainder of the paper is organized as follows. GMR and GPR are reviewed and compared in Section~\ref{sec:Background}. The proposed GMR-based GP is introduced in Section~\ref{sec:GMRbasedGP} and validated in a real-robot experiment in Section~\ref{sec:Experiments}. Finally, Section~\ref{sec:Discussion} presents a discussion on similarities and differences of the proposed approach compared to other probabilistic methods, notably ProMP and KMP. 

\section{Background}
\label{sec:Background}
\subsection{Gaussian mixture regression}
\label{subSec:GMR}

Gaussian mixture regression (GMR) exploits the Gaussian conditioning theorem to estimate the distribution of output data given input data \citep{Ghahramani1994,Ghahramani1996}. A Gaussian mixture model (GMM) is first estimated to encode the joint distribution of input and output datapoints, e.g., with an Expectation-Maximization (EM) algorithm. The output given observed input is then predicted via a linear combination of conditional expectations. Hence, GMR does not fit the regression function directly, but relies instead on the learned joint distribution.  

We denote $\bm{X}$ and $\bm{Y}$ random vectors of input and corresponding output data, respectively, and by $\bm{x}$ and $\bm{y}$ arbitrary realizations of them. In a GMM with $C$ components, the joint distribution of $(\bm{X}^\trsp,\bm{Y}^\trsp)^\trsp$ is encoded by  
\begin{equation}
\Big( 
\begin{smallmatrix}
\bm{X}_{}^{} \\
\bm{Y}_{}^{} \\
\end{smallmatrix} \Big)
\sim \sum_{\indCl=1}^{C} \pi_\indCl\; 
\mathcal{N} \bigg( 
\Big(\begin{smallmatrix}
\bm{\mu}_\indCl^{\ty{X}} \\
\bm{\mu}_\indCl^{\ty{Y}} \\
\end{smallmatrix} \Big)
,\Big( \begin{smallmatrix} \bm{\Sigma}_\indCl^{\ty{X}} &
\bm{\Sigma}_\indCl^{\ty{XY}} \\ 
\bm{\Sigma}_\indCl^{\ty{YX}} &
\bm{\Sigma}_\indCl^{\ty{Y}}\\ \end{smallmatrix} \Big)
\bigg),
\label{Eq:GMM}
\end{equation}
with 
$\pi_\indCl$, $\bm{\mu}_\indCl$ and $\bm{\Sigma}_\indCl$ the mixing coefficient (prior), mean and covariance matrix of the $\indCl$-th component. 

GMR computes the conditional distribution of the GMM joint distribution to infer the output vector corresponding to a given input vector. The resulting multimodal distribution possesses second order moments that can be calculated from the conditional means and covariances of the multivariate Gaussian distributions associated with individual components using the laws of total mean and covariance, so that
\begin{equation}
\bm{\hat{y}}^{\text{M}}(\bm{x}) =
\sum\limits_{\indCl=1}^{C}
h_\indCl (\bm{x}) 
\bm{\hat{y}}_\indCl (\bm{x}) \;\text{ and }\;
\bm{\hat{\Sigma}}^{\text{M}} (\bm{x})  = 
\sum\limits_{\indCl=1}^{C} 
h_\indCl (\bm{x})
\bm{\tilde{\Sigma}}_{\ell}(\bm{x}) 
-\; \bm{\hat{y}}^{\text{M}}(\bm{x}) 
(\bm{\hat{y}}^{\text{M}} (\bm{x}) )^\trsp, \label{Eq:GMR}
\end{equation}
with componentwise conditional means and covariances
\begin{equation*}
\bm{\hat{y}}_\indCl (\bm{x})
= \bm{\mu}_\indCl^{\ty{Y}} + 
\bm{\Sigma}_\indCl^{\ty{YX}}
{\bm{\Sigma}_\indCl^{\ty{X}}}^{-1}
\left(\bm{x}-\bm{\mu}_\indCl^{\ty{X}}\right) \;\text{ and }\;
\bm{\hat{\Sigma}}_\indCl = 
\bm{\Sigma}_\indCl^{\ty{Y}} - 
\bm{\Sigma}_\indCl^{\ty{YX}}
{\bm{\Sigma}_\indCl^{\ty{X}}}^{-1}
\bm{\Sigma}_\indCl^{\ty{XY}},
\end{equation*}
and 
$ \bm{\tilde{\Sigma}}_{\ell}(\bm{x})=\bm{\hat{\Sigma}}_\indCl
+\bm{\hat{y}}_\indCl (\bm{x})  
(\bm{\hat{y}}_\indCl (\bm{x}))^\trsp$. 
The so-called responsability $h_\indCl$ of component $\indCl$ are computed in closed form as 
\begin{equation}
h_\indCl (\bm{x}) = 
\frac{\pi_\indCl\, \phi\left(
	\bm{x};\bm{\mu}_\indCl^{\ty{X}},\bm{\Sigma}_\indCl^{\ty{X}}
	\right)}
{\sum\limits_{i=1}^{C}\pi_i\, 
	\phi\left(
	\bm{x};\bm{\mu}_i^{\ty{X}},\bm{\Sigma}_i^{\ty{X}}
	\right)},
\label{Eq:GMR_resp}
\end{equation}
where $\phi\left(\bm{x};\bm{\mu},\bm{\Sigma}\right)$ stands for the probability density function at point $\bm{x}$ of the multivariate Gaussian distribution with mean $\bm{\mu}$ and covariance matrix $\bm{\Sigma}$.
The computational complexity of GMR is mainly dependent on the number of GMM components as it governs the dimensionality of the maximum likelihood problem usually tackled by Expectation-Maximization. Moreover, the number of GMM components is the only parameter that needs to be specified and can be estimated online, e.g., with a Bayesian nonparametric approach \citep{Tanwani19}.
Therefore, GMR is well adapted for real-time application and its simplicity allows it to be combined easily to other complementary approaches.

Figure~\ref{subFig:BaseMethodsGMR} shows an example of application of GMR. The training dataset consists of 5 demonstrations of a two-dimensional time-driven trajectory. A GMM ($C=6$) is first trained to encode the joint distribution of the inputs $t$ and outputs $\bm{y}$. The conditional distribution of $\bm{y}$ given $t$ is inferred by GMR. The covariance $\bm{\hat{\Sigma}}^{\text{M}} (\bm{x})$ of the distribution encodes the variability of the demonstrations. The output distribution is extrapolated in the absence of training data ($t>2$). 

\subsection{Gaussian process regression}
\label{subSec:GPR}

Gaussian processes (GPs) form a class of probabilistic models that aims at learning a deterministic input-output relationship, up to observation noise, based on a Gaussian prior over potential objective functions. In the noiseless GP framework, the output $y$ is hence typically seen as a function of a controlled input $\bm{x}$. Randomness comes in an instrumental way as the function $y(\bm{x})$ is assumed to be one realization of a Gaussian random process or random function denoted $Y(\bm{x})$. 
Predictions of the objective function are then made by relying on the conditional distribution of $Y(\bm{x})$ knowing that $Y(\bm{x}_\indData^{(\text{o})})$ coincides with the observed outputs $\bm{y}_\indData^{(\text{o})}$ at their corresponding observation inputs $\bm{x}_\indData^{(\text{o})}$. 

Multi-output Gaussian processes (MOGPs) generalize GPs to vector-valued output by predicting jointly the output components (see \citep{Alvarez2012} for a review). Therefore, MOGPs exploits the potential relation between the output components, which are not taken into account if predictions are computed separately for each dimension.
Similarly to standard GP, the vector-valued objective function is modeled in MOGP with a vector-valued GP $(\bm{Y}(\bm{x}))$, inducing finite-dimensional prior distributions $\bm{Y}(\bm{x}_{1:N}) \; \sim \; \mathcal{N}\left(\bm{\mu}_{\bm{x}_{1:N}}, \bm{K}_{\bm{x}_{1:N}} \right)$ for any arbitrary set of inputs $\bm{x}_{1:N}$. We here denote $\bm{\mu}_{\bm{x}_{1:N}} = \bm{\mu}(\bm{x}_{1:N})$ and $\bm{K}_{\bm{x}_{1:N}} = \bm{k}(\bm{x}_{1:N},\bm{x}_{1:N})$ the corresponding mean vector and covariance matrix, where $\bm{\mu}$ and $\bm{k}$ stand for the mean function and cross-covariance kernel of the MOGP and $\bm{\mu}_{\bm{x}}\in\mathbb{R}^{D}$, $\bm{K}_{\bm{x}}\in\mathbb{R}^{D\times D}$ with $D$ the output dimension.

\begin{figure}[tbp]
	\centering
	\begin{subfigure}[b]{0.45\textwidth}
		\begin{minipage}{0.57\textwidth}
			\centering
			\includegraphics[width=\textwidth]{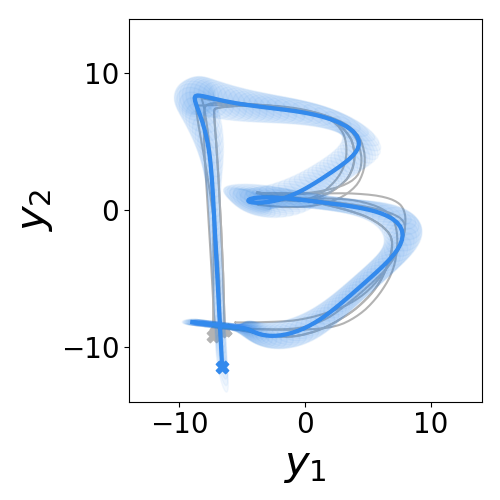}
		\end{minipage}
		\begin{minipage}{0.4\textwidth}
			\includegraphics[width=\textwidth]{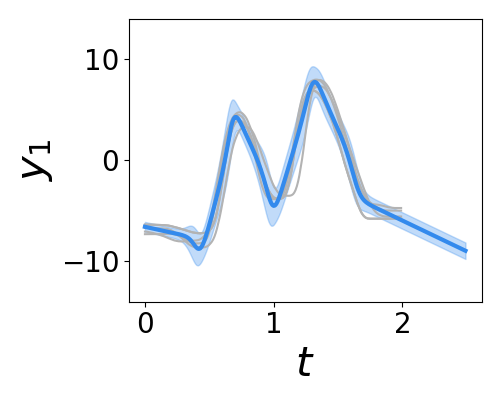}
			\includegraphics[width=\textwidth]{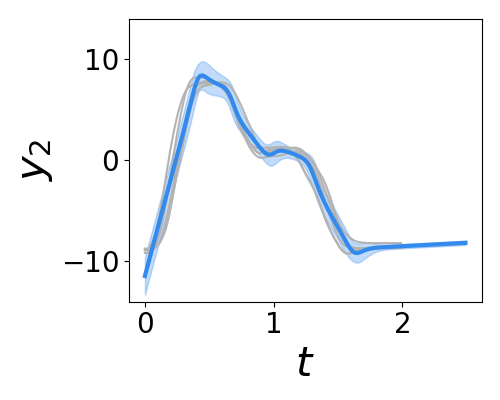}	
		\end{minipage}
		\caption{}
		\label{subFig:BaseMethodsGMR}
	\end{subfigure}
	\begin{subfigure}[b]{0.45\textwidth}
		\begin{minipage}{0.57\textwidth}
			\centering
			\includegraphics[width=\textwidth]{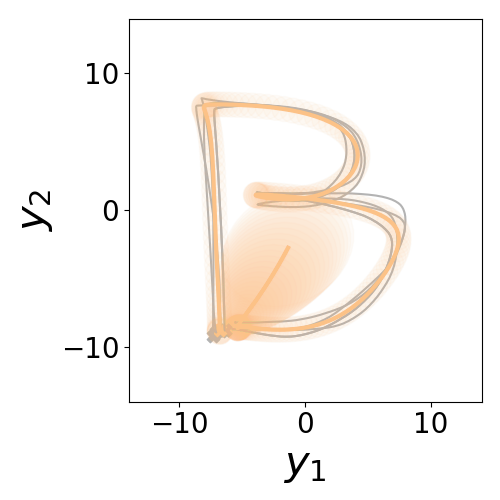}
		\end{minipage}
		\begin{minipage}{0.4\textwidth}
			\includegraphics[width=\textwidth]{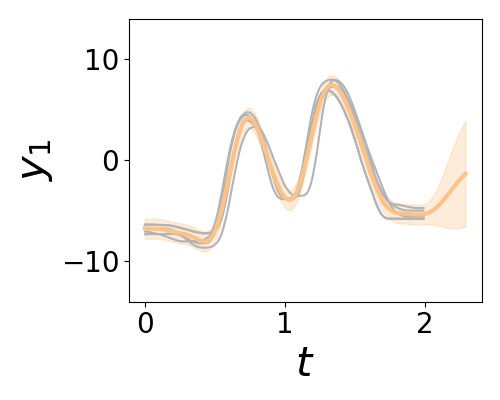}
			\includegraphics[width=\textwidth]{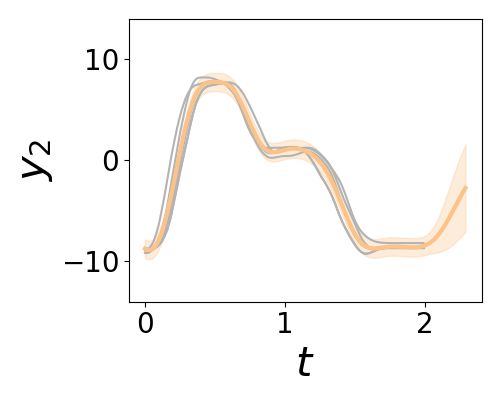}	
		\end{minipage}
		\caption{}
		\label{subFig:BaseMethodsGPR}
	\end{subfigure}
	\caption{Comparison between (\emph{a}) GMR and (\emph{b}) GPR. The training data are shown in light gray for all graphs. The mean is represented by a continuous line and the variance by a light tube around the estimate. The \emph{left} graphs show the output trajectories estimated by GMR and GPR, respectively. The beginning of the trajectories is marked by a cross. The \emph{right} graphs show the estimated trajectory for each output component as a function of the input $t$. Time and positions are given in seconds and centimeters, respectively.} 
	\label{Fig:BaseMethods}
\end{figure}

Denoting $\bm{y}^{(\text{o})}_{1:N}$ the observed realization of $\bm{Y}(\bm{x}_{1:N}^{(\text{o})})+ \boldsymbol{\varepsilon}$, the posterior distribution follows by Gaussian conditioning
\begin{equation}
\bm{Y}(\bm{x})|
\bm{y}^{(\text{o})}_{1:N} \;\sim\; 
\mathcal{N}(
\bm{\hat{y}}^{\text{P}}(\bm{x}),\bm{\hat{\Sigma}}^{\text{P}}
),
\label{Eq:GPR_posterior}
\end{equation}
with conditional mean and covariance functions given by
\begin{equation}
\bm{\hat{y}}^{\text{P}}(\bm{x}) = 
\bm{\mu}_{\bm{x}} + 
\bm{K}_{\bm{x},\bm{x}_{1:N}^{(\text{o})}} 
\left( \bm{K}_{\bm{x}_{1:N}^{(\text{o})}} + \bm{\Sigma}_{\boldsymbol{\varepsilon}}  \right)^{-1} 
(\bm{y}^{(\text{o})}_{1:N}-\bm{\mu}_{\bm{x}_{1:N}^{(\text{o})}}),
\label{Eq:GPR_mean}
\end{equation}
\begin{equation}
\bm{\hat{\Sigma}}^{\text{P}}(\bm{x}) =
\bm{K}_{\bm{x}} - 
\bm{K}_{\bm{x},\bm{x}_{1:N}^{(\text{o})}}
\left( \bm{K}_{\bm{x}_{1:N}^{(\text{o})}} + \bm{\Sigma}_{\boldsymbol{\varepsilon}}  \right)^{-1} 
\bm{K}_{\bm{x}_{1:N}^{(\text{o})}, \bm{x}},
\label{Eq:GPR_covariance}
\end{equation}
where $\bm{\Sigma}_{\boldsymbol{\varepsilon}}$ is the covariance matrix of the observation noise $\boldsymbol{\varepsilon}$, which is assumed centered Gaussian and independent of the process $\bm{Y}$.
The covariance $\bm{\hat{\Sigma}}^{\text{P}}(\bm{x})$ expresses the prediction uncertainty for all components and between them. 
In typical cases, the further away the input data lies from the training dataset the larger the prediction variance, as illustrated in Fig.~\ref{Fig:BaseMethods}-(\emph{right}). Moreover, the mean $\bm{\hat{y}}^{\text{P}}(\bm{x})$ then converges to the prior mean $\bm{\mu}_{\bm{x}}$, equal to $\bm{0}$ in this case.

The class of covariance kernels that we consider in this paper is formulated as a sum of separable kernel functions generated under the linear model of coregionalization (LMC) assumption \citep{Goovaert1997}. This class of kernel functions is often called separable as the dependencies between inputs and outputs are decoupled. Therefore, the kernel $\bm{k}(\bm{x},\bm{x}')$ between two input vectors $\bm{x}$ and $\bm{x}'$ is expressed as
\begin{align}
\bm{k}\left( \bm{x}, \bm{x}' \right) 
= \sum\limits_{q=1}^{Q} \bm{\Upsilon}_q k_q\left( \bm{x}, \bm{x}' \right),
\label{Eq:GPR_sumSepKernels}
\end{align}
where the so-called coregionalization matrices $\bm{\Upsilon}_q\in\mathbb{R}^{D\times D}$ are positive semi-definite matrices representing the interaction among the output components.
The choice of the scalar-valued kernels $k_q$ and the design of the coregionalization matrices $\bm{\Upsilon}_q$ are crucial for the GP as they represent our prior knowledge about the function that is being learned. 

Figure~\ref{subFig:BaseMethodsGPR} shows an example of MOGP with the separable kernel \eqref{Eq:GPR_sumSepKernels} where $Q=1$ and $k_q\left( \bm{x}, \bm{x}' \right)$ is a Mat\'ern kernel ($\nu=5/2$), so that
$k_q\left( \bm{x}, \bm{x}' \right) = \sigma_f^2 \Big(1 + \frac{\sqrt{5}r}{\sigma_l} + \frac{5r^2}{3\sigma_l^2}\Big)\exp\Big(-\frac{\sqrt{5}r}{\sigma_l}\Big),$
where $r=\sqrt{(\bm{x}-\bm{x}')^\trsp(\bm{x}-\bm{x}')}$ is the Euclidean distance between inputs and $\sigma_f$, $\sigma_l$ are the variance and lengthscale parameters, respectively.

\section{GMR-based Gaussian Processes}
\label{sec:GMRbasedGP}
In this section, we propose to combine GMR and GPR to form a GMR-based GP. The proposed approach takes advantage of the ability of GPs to encode various prior beliefs through the mean and kernel functions and allows the variability information retrieved by GMR to be encapsulated in the uncertainty estimated by the GP. Moreover, the proposed approach enjoys the properties of generative models, therefore new trajectories can be easily generated through sampling and conditioning.

\subsection{GMR-based GPs formulation}
\label{subSec:GMRbasedGPR}

We define the GMR-based GP as a GP with prior mean
\begin{equation}
\bm{\mu}(\bm{x}) = \bm{\hat{y}}^{\text{M}}(\bm{x}),
\label{Eq:GMRbasedGPmean}
\end{equation}
and a kernel in the form of a sum of $C$ separable kernels associated with the $C$ components of the considered GMM
\begin{equation}
\bm{k}(\bm{x},\bm{x}') = 
\sum\limits_{\indCl=1}^{C} 
h_\indCl(\bm{x}) h_\indCl(\bm{x}')
\bm{\hat{\Sigma}}_\indCl
k_\indCl(\bm{x},\bm{x}').
\label{Eq:GMRbasedGPkernel}
\end{equation}
The prior mean of Eq.~\eqref{Eq:GMR} allows the GP to follow the GMR predictions far from training data.
Moreover, the constructed GP is also covariance non-stationary due to its spatially-varying coregionalization matrices \citep{Gelfand04}. The input-dependent coregionalization matrices $h_\indCl(\bm{x}_m) h_\indCl(\bm{x}_n)\bm{\hat{\Sigma}}_\indCl$ corresponding to this conception are determined by GMR (via \eqref{Eq:GMR}, \eqref{Eq:GMR_resp}). Alternatively, one can say that the GMR responsabilities $h_\indCl$ weight the importance of the kernels $k_\indCl(\bm{x}_m,\bm{x}_n)$ according to the proximity of the input data to the center of GMM components. Thus, the kernels associated to the centers close to the given input data are more relevant than distant centers. The covariance matrices $\bm{\hat{\Sigma}}_\indCl$ allows the dependencies between the output data to be described for each GMM component. Note that both the coregionalization matrices and the number of separable kernels are determined by GMR. Therefore, the only parameters to determine are the hyperparameters of the kernels $k_\indCl$ which can be estimated, for example, by maximizing the likelihood of the GP. Moreover, the variance parameters $\sigma_f$ of the kernels $k_\indCl$ are fixed  to $1$ as they are already scaled by the covariance matrices $\bm{\hat{\Sigma}}_\indCl$. Thus, the estimation of hyperparameters is simplified compared to standard LMC.

\begin{figure}[tbp]
	\centering 
	\begin{subfigure}[b]{0.45\textwidth}
		\begin{minipage}{0.57\textwidth}
			\includegraphics[width=\textwidth]{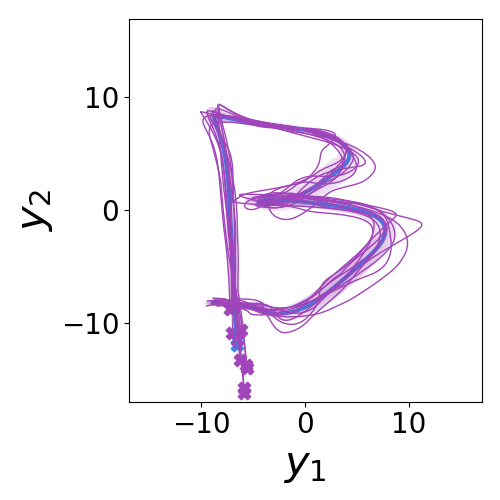}
		\end{minipage}
		\begin{minipage}{0.4\textwidth}
			\includegraphics[width=\textwidth]{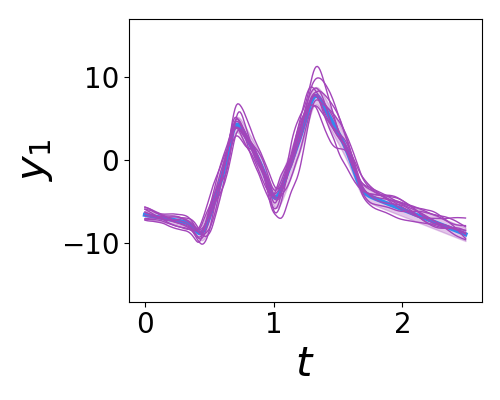}
			\includegraphics[width=\textwidth]{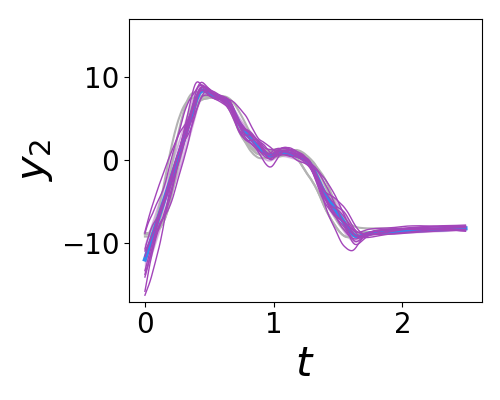}
		\end{minipage}
		\caption{}
		\label{Fig:GMRbasedGPRprior}
	\end{subfigure}
	\begin{subfigure}[b]{0.45\textwidth}
		\begin{minipage}{0.57\textwidth}
			\includegraphics[width=\textwidth]{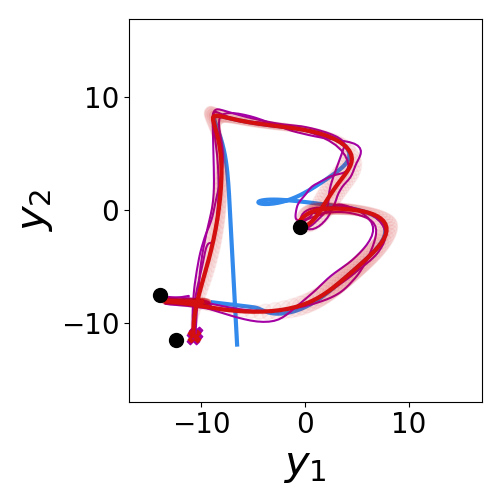}
		\end{minipage}
		\begin{minipage}{0.4\textwidth}
			\includegraphics[width=\textwidth]{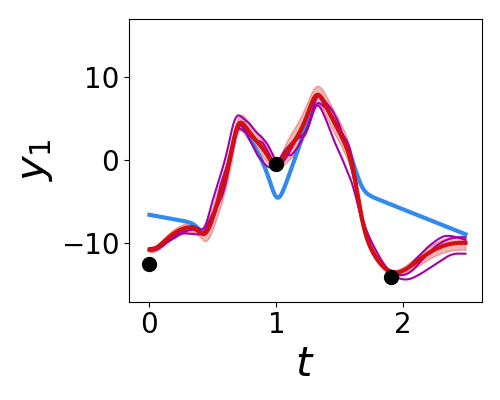}
			\includegraphics[width=\textwidth]{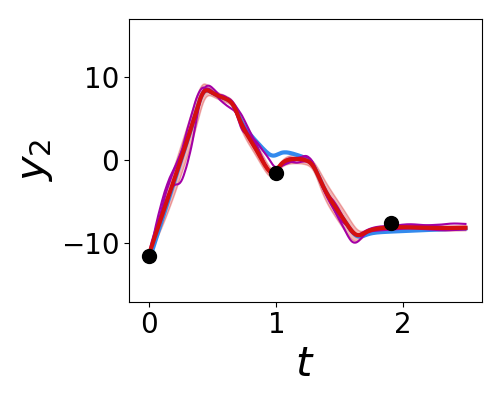}
		\end{minipage}
		\caption{}
		\label{Fig:GMRbasedGPRposterior}
	\end{subfigure}
	\caption{(\emph{a}) Sample trajectories generated from GMR-based GP. The prior mean of the process and the sample trajectories are represented by continuous blue and purple lines, respectively. The covariance $\bm{K}_{t_{1:N}^{(\text{o})}}$ of the process is represented by a light purple tube around the prior mean. (\emph{b}) Sample and predicted trajectories generated from the posterior distribution of the GMR-based GP. The prior, sampled and predicted trajectories are represented by continuous blue, dark pink and red lines, respectively. The uncertainty of the prediction is represented by a light red tube around the predicted mean. Via-points, considered as observations for the GMR-based GP, are represented by black dots. The trajectories are extrapolated from training data for $t>2$.}
	\label{Fig:GMRbasedGPR}  	
\end{figure}

Figure~\ref{Fig:GMRbasedGPRprior} shows the prior mean and 10 sample trajectories generated from the proposed GMR-based GP where $k_\indCl$ are Mat\'ern kernels ($\nu=5/2$). The hyperparameters, namely the lengthscales $\sigma_{l,\indCl}$ of the $k_\indCl$ and the covariance of the noise $\bm{\Sigma}_{\bm{\epsilon}} = \sigma_\epsilon \bm{I}$, were optimized by maximum likelihood estimation within the GPy framework \citep{gpy2014}. Note that the prior mean of the process corresponds to the mean obtained by GMR in Figure~\ref{subFig:BaseMethodsGMR}. Moreover, the prior uncertainty provided by the GMR-based GP is lower in the regions where the variability of the demonstrations is low, e.g. at the bottom of the straight vertical line of the \emph{B} letter, and higher in the regions of higher variability, e.g. in the curves in the right part of the \emph{B}.

\subsection{GMR-based GPs properties}
\begin{figure}[!tbp]
	\centering
	\begin{subfigure}[b]{0.19\textwidth}
		\centering
		\includegraphics[width=\textwidth]{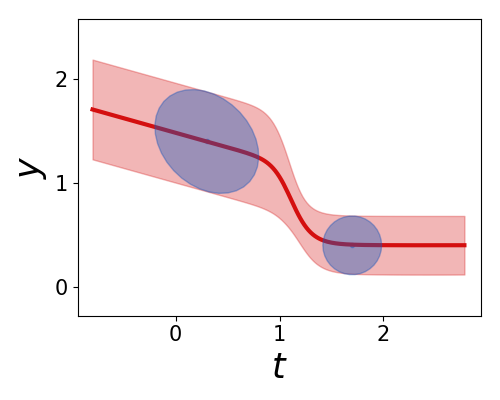}
		\caption{}
		\label{subFig:GmrBasedGPR_toyEx_0obs}
	\end{subfigure}
	\begin{subfigure}[b]{0.19\textwidth}
		\centering
		\includegraphics[width=\textwidth]{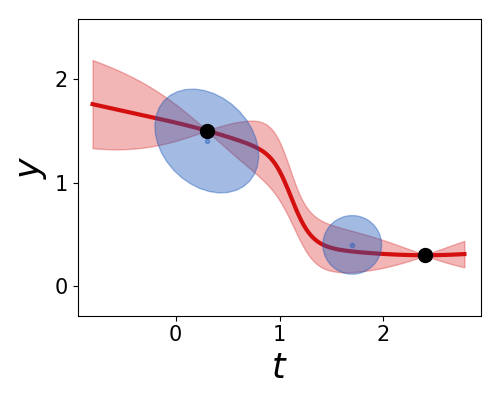}
		\caption{}
		\label{subFig:GmrBasedGPR_toyEx_2obs}
	\end{subfigure}
	\begin{subfigure}[b]{0.19\textwidth}
		\centering
		\includegraphics[width=\textwidth]{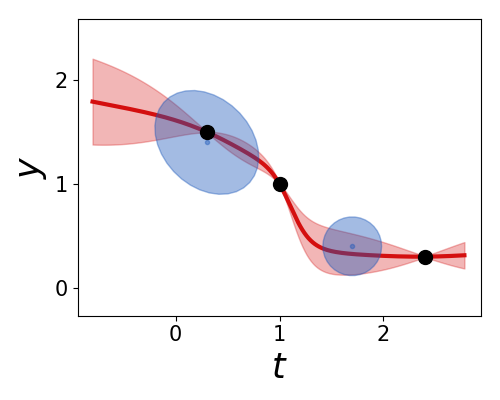}
		\caption{}
		\label{subFig:GmrBasedGPR_toyEx_3obs}
	\end{subfigure}
	\begin{subfigure}[b]{0.19\textwidth}
		\centering
		\includegraphics[width=\textwidth]{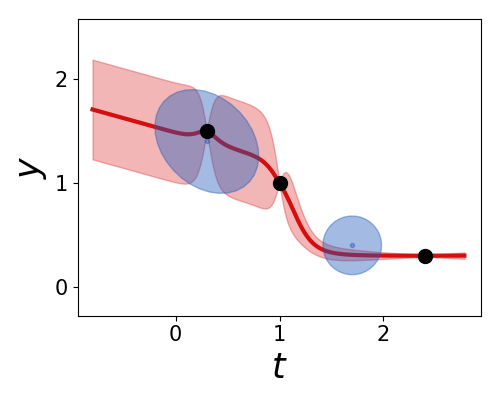}
		\caption{}
		\label{subFig:GmrBasedGPR_toyEx_lengthscale02}
	\end{subfigure}
	\begin{subfigure}[b]{0.19\textwidth}
		\centering
		\includegraphics[width=\textwidth]{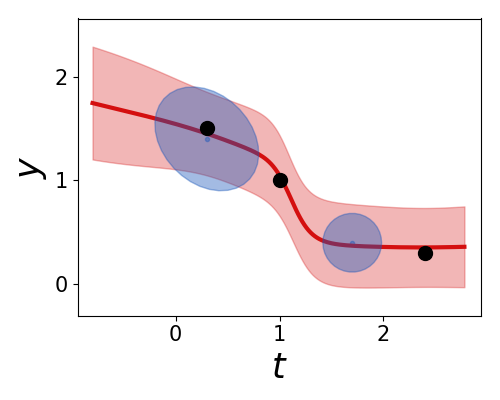}
		\caption{}
		\label{subFig:GmrBasedGPR_toyEx_noise}
	\end{subfigure}
	\caption{Example of time-driven 1D-trajectories predicted by GMR-based GP. The process was constructed based on a 2-components GMM, represented by blue ellipses. The estimate of GMR-based GPR is shown by a red continuous line with the corresponding uncertainty represented by a light tube around the mean. The initial observations are discarded after determining the hyperparameters. The lenghtscale parameters and noise covariance are fixed as $\sigma_l= 1$ and $\sigma_\epsilon = 1e^{-4}$, respectively. (\emph{a}) The posterior distribution without any new observation is represented. 
	(\emph{b}-\emph{c}) Two, respectively 3, via-points, represented with black dots, are added as observations of the GMR-based GP. 
	(\emph{d}) The lengthscale parameters are fixed as $\sigma_l=0.1$, $\sigma_l=5$ for the left and right GMM component, respectively.
	(\emph{e}) The noise covariance of the process is fixed as $\sigma_\epsilon=0.1$ ($\sigma_l=1$).}
	\label{Fig:GmrBasedGPR_toyEx}
\end{figure}

A particularity of the presented GMR-based GP is that the information on the demonstrations distribution is included in the prior mean $\bm{\mu}_{\bm{x}}$ and covariance $\bm{K}_{\bm{x}}$ after determining the hyperparameters. Therefore, the training data can be ignored and our model can be conditioned uniquely on new observations.
Figure~\ref{subFig:GmrBasedGPR_toyEx_0obs} shows the mean and uncertainty recovered by a 1D-output GMR-based GP without any observation. The process was constructed based on a GMM with two components with $k_\indCl$ defined as Mat\'ern kernels ($\nu=5/2$). The lengthscale parameters $\sigma_l$ of the $k_\indCl$ and the covariance of the noise of the process $\sigma_{\epsilon}$ are fixed as equal to $1$ and $1e^{-4}$, respectively. Note that the distribution corresponds to the prior of the GMR-based GP, therefore the mean is exactly equal to the mean computed by GMR. Moreover, if a component $\indCl$ is entirely responsible for a test datapoint so that $h_\indCl=1$, the corresponding uncertainty is equal to the conditional covariance of the component $\bm{\hat{\Sigma}}_\indCl$ augmented with $\Sigma_{\epsilon}$, as observed for $t<0.6$ and $t>1.8$ in Figure~\ref{subFig:GmrBasedGPR_toyEx_0obs}. In the case where several components are responsible for the datapoint, its uncertainty is a weighted sum of the conditional covariances, as observed for the zone in between the two GMM components. Therefore, the prior uncertainty obtained by GMR-based GP without observation reflects the variability provided by GMR. However, note that the prior uncertainty of GMR-based GP is not equal to $\bm{\hat{\Sigma}}^{\text{M}}$.

In the cases where it is desirable to adapt trajectories towards new start-, via- or end-points $(\bm{\xi}_v, \bm{\zeta}_v)$, those particular points are used to define a new set of observation inputs and outputs $(\bm{x}_{1:V}^{(o)}, \bm{y}_{1:V}^{(o)}) = (\bm{\xi}_{1:V}, \bm{\zeta}_{1:V})$ which is then used to infer the posterior distribution of the GMR-based GP with \eqref{Eq:GPR_posterior}. Figures~\ref{subFig:GmrBasedGPR_toyEx_2obs} and~\ref{subFig:GmrBasedGPR_toyEx_3obs} show examples where 2 and 3, via-points were added to the trajectory. We observe that the mean of the process goes through the via-points and the uncertainty becomes very small in these locations. Note that conditioning a trajectory towards via-points with GMR alone is not straightforward due to the fact that covariance terms between two datapoints are equal to zero.

As in a standard GP, the predicted mean and uncertainty depend strongly on the kernel parameters. 
Moreover, one of the advantages of the GMR-based GP is that each kernel $k_\indCl$ can be chosen individually and their parameters are determined separately. Therefore, different behaviors can be obtained as a function of the location in the input space, as shown by Figure~\ref{subFig:GmrBasedGPR_toyEx_lengthscale02} where the lengthscale parameters of the kernels corresponding to the left and right GMM component are equal to $0.1$ and $5$, respectively. Similarly to a standard GP, the noise of the process determines the behavior of the GMR-based GP at the via-points location. As shown by Figure~\ref{subFig:GmrBasedGPR_toyEx_noise} where $\sigma_\epsilon=0.1$, the constraint of passing exactly through the via-points is alleviated and the mean of the GMR-based GP passes close-by the via-points while the uncertainty is equal to the noise of the process. Note that the noise parameter can also be defined separately for each kernel $k_\indCl$.

Figure~\ref{Fig:GMRbasedGPRposterior} shows the predicted mean and corresponding uncertainty as well as three trajectories sampled from the posterior distribution of the GMR-based GP on the \emph{B} trajectory. As explained previously, the initial demonstrations data were used for hyperparameters estimation but not incorporated as conditioning data and three via-points have been added as new observations. We observe that the estimate and the posterior trajectories are adapted to pass close to the via-points. In the zones far from the via-points, the predicted trajectory follows the prior mean of the process.

\section{Experiments}
\label{sec:Experiments}

\begin{figure}[tbp]
	\centering 
	\begin{subfigure}[b]{0.63\textwidth}
		\centering
		\includegraphics[width=.32\textwidth]{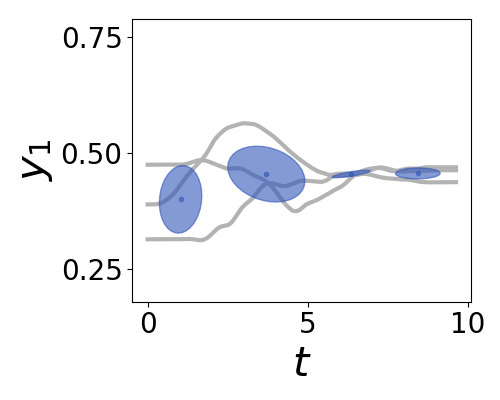}
		\includegraphics[width=.32\textwidth]{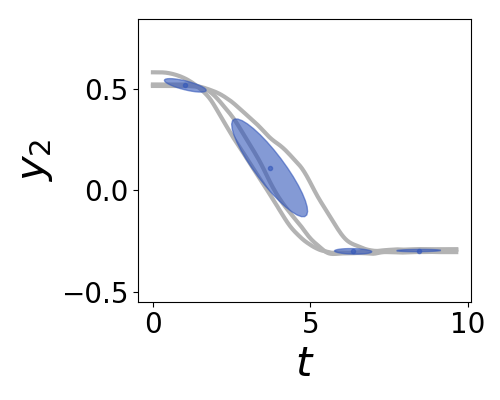}
		\includegraphics[width=.32\textwidth]{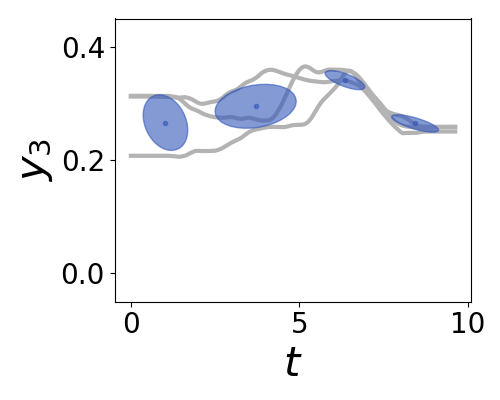}
		\includegraphics[width=.32\textwidth]{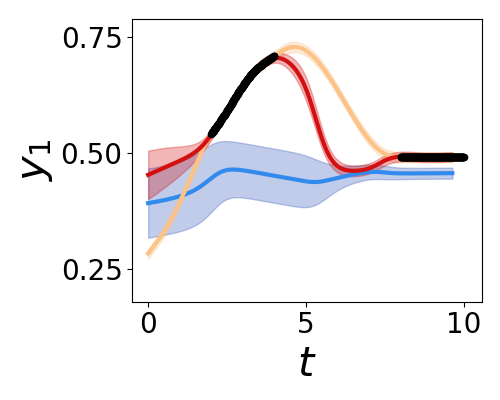}
		\includegraphics[width=.32\textwidth]{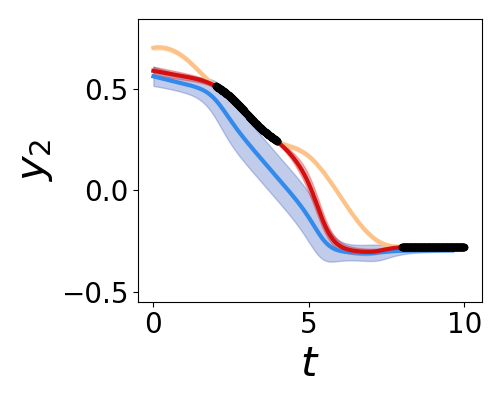}
		\includegraphics[width=.32\textwidth]{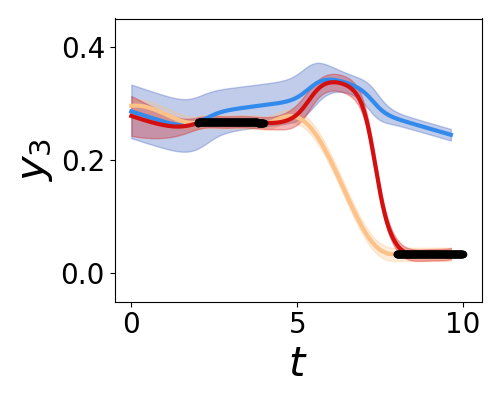}
		\caption{}
		\label{subFig:Exp_gmm_repros}
	\end{subfigure}
	\begin{subfigure}[b]{0.35\textwidth}
		\centering
		\includegraphics[width=\textwidth]{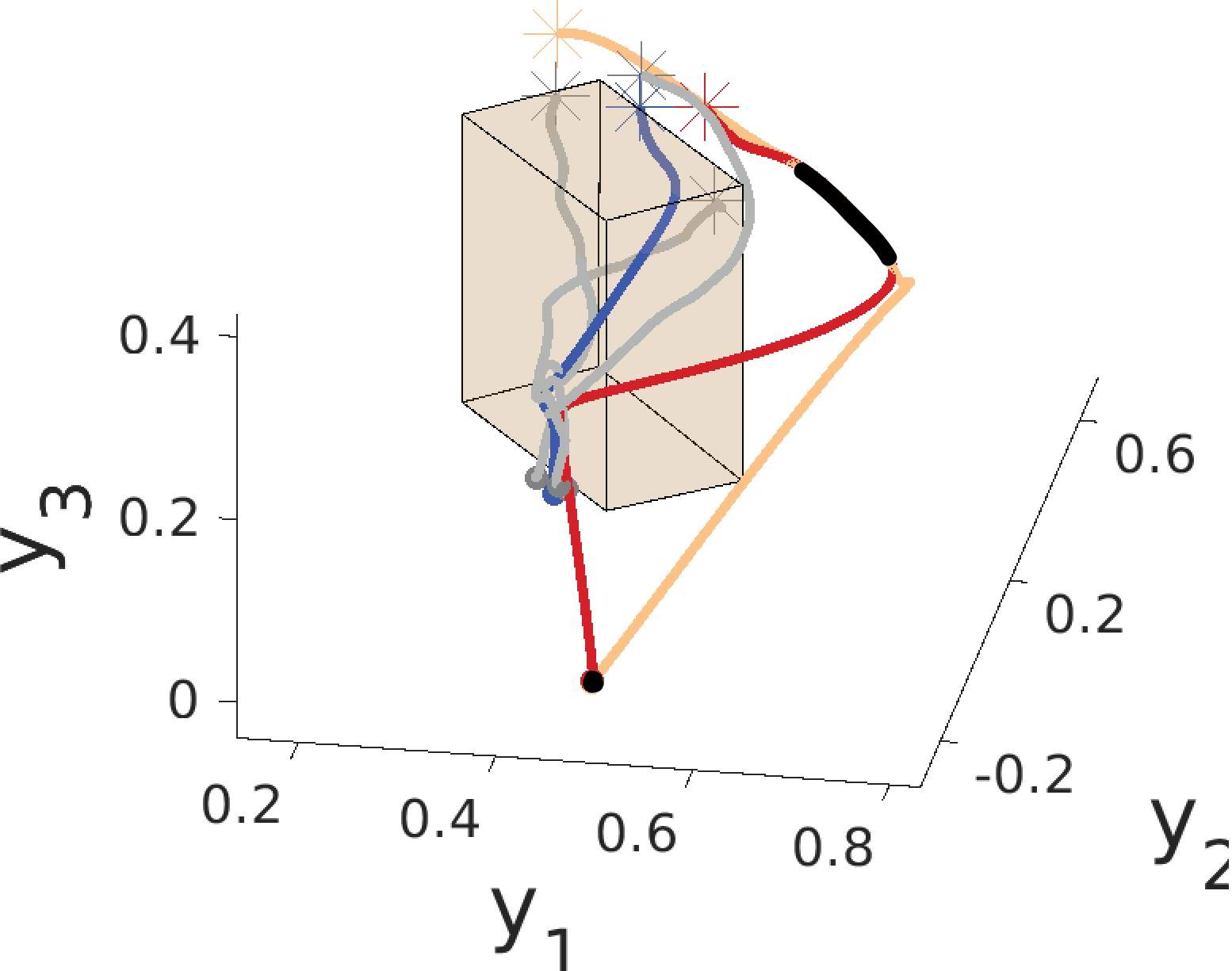}
		\caption{}
		\label{Fig:Exp_repro3d}
	\end{subfigure}
	\caption{(\emph{a})-\emph{top} Demonstrated trajectories (in light gray) and corresponding GMM ($K=4$) represented as blue ellipses. The Cartesian positions $y_1$, $y_2$, $y_3$, considered as outputs, are represented as a function of the time, considered as input.  (\emph{a})-\emph{bottom} Reproduced trajectories. The means of the trajectories generated by GMR-based GP, MOGP and GMR are represented by red, yellow and blue lines, respectively, with their respective covariance depicted by light tubes around the estimates. 
	The via-points defined to modulate the trajectories generated by GMR-based GP and MOGP are depicted with black dots. (\emph{b}) 3D representation of the reproduced trajectories by the robot. The beginnings of the demonstrated and reproduced trajectories are depicted by stars. 
	}  
	\label{Fig:Exp_repro_traj}
\end{figure}

In this section, we evaluate the proposed approach in a peg-in-hole task achieved by the 7-DoF Franka Emika Panda robot. In the first part of the experiment, 3 demonstrations of the task were collected by kinesthetically guiding the robot to first approach a hollow cylinder and then insert the peg in it. For all the demonstrations, the hollow cylinder was placed 20 cm above the table. The collected data, encoding time $t$ and Cartesian position $(y_1 \; y_2 \; y_3)^\trsp$, were time-aligned. We trained a GMM and determined the hyperparameters of a GMR-based GP, as well as a MOGP with the separable kernel \eqref{Eq:GPR_sumSepKernels} ($Q=C$) based on the time-driven demonstrated trajectories. Mat\'ern kernels ($\nu=5/2$) were chosen as individual kernels $k_\indCl$ and $k_q$ for the GMR-based GP and MOGP, respectively. The number of components of the GMM ($C=4$) was selected by the experimenter. Figure~\ref{subFig:Exp_gmm_repros} shows the demonstrated trajectories and corresponding GMM states.

In the second part of the experiment, an obstacle was added in between the initial position of the robot and the hollow cylinder. Moreover, the hollow cylinder was positioned directly on the table, i.e. 20 cm below its location during the demonstrations. Via-points were determined by the experimenter to modulate trajectories so that the robot avoids the obstacle in the desired manner and its final position corresponds to the new location of the hollow cylinder. The performances of the proposed GMR-based GP, the MOGP and GMR to reproduce the task in the modified environment were compared.

As explained in the previous section, the via-points were used to define a new set of observations for the GMR-based GP, while the original training data are discarded after inferring the hyperparameters. In the case of the MOGP, the mean and uncertainty of the reproduction considering $V$ via-points are updated for each testing input $\bm{x}$ by conditioning the distribution \eqref{Eq:GPR_posterior} on the desired via-points.
The mean and variability of the reproduction obtained by GMR can be updated in a similar way. However, as the covariances between different datapoints are not encoded in GMR, the generated trajectory is discontinuous. Therefore, we did not reproduce it with the robot and we show instead in the following graphs the GMR reproduction where no via-points are considered, whose mean corresponds to the prior mean of the GMR-based GP.

The task was reproduced using a linear quadratic regulator (LQR) controller tracking the trajectory predicted by GMR-based GPR, MOGP or GMR \citep{Calinon2016}. The required tracking precision was set as proportional to the inverse of the posterior covariance $\bm{\hat{\Sigma}}$ of the different methods. This information is exploited to demand a high precision tracking in the regions of the trajectories where high certainty (GMR-based GP and MOGP) or low variability (GMR) are observed, and vice-versa.

\begin{figure*}[tbp]
	\centering 
	\begin{subfigure}[b]{\textwidth}
		\centering
		\includegraphics[width=0.16\textwidth]{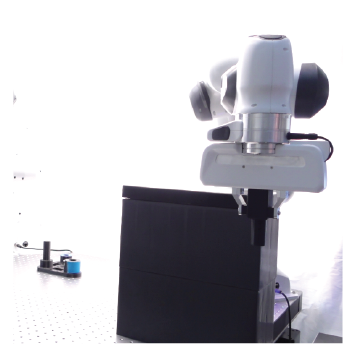}
		\includegraphics[width=0.16\textwidth]{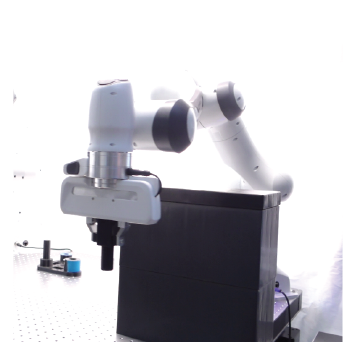}
		\includegraphics[width=0.16\textwidth]{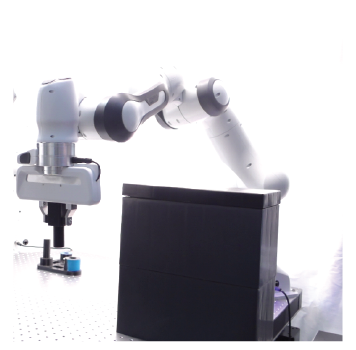}
		\includegraphics[width=0.16\textwidth]{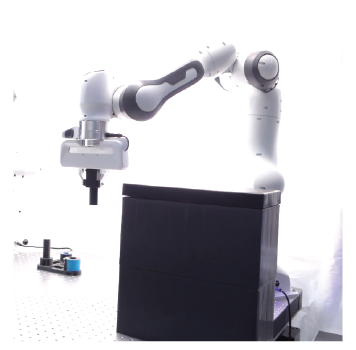}
		\includegraphics[width=0.16\textwidth]{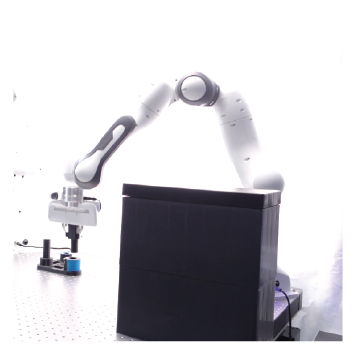}
		\includegraphics[width=0.16\textwidth]{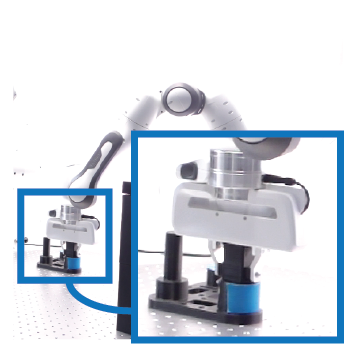}
		\label{subFig:Exp_reproGMRbGP}
	\end{subfigure}
	\begin{subfigure}[b]{\textwidth}
		\centering
		\includegraphics[width=0.16\textwidth]{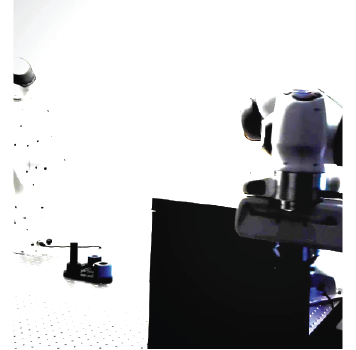}
		\includegraphics[width=0.16\textwidth]{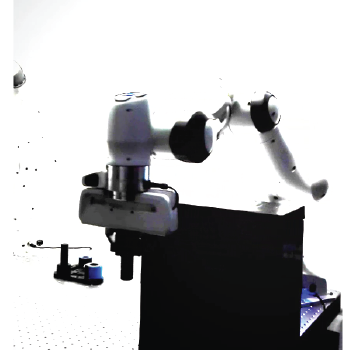}
		\includegraphics[width=0.16\textwidth]{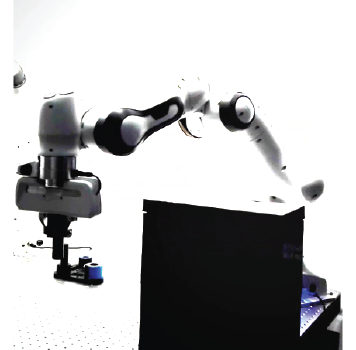}
		\includegraphics[width=0.16\textwidth]{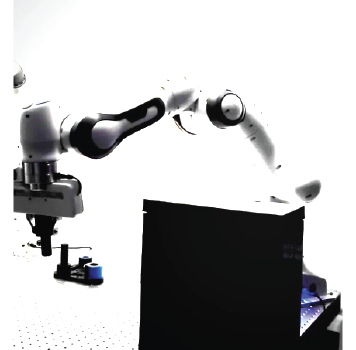}
		\includegraphics[width=0.16\textwidth]{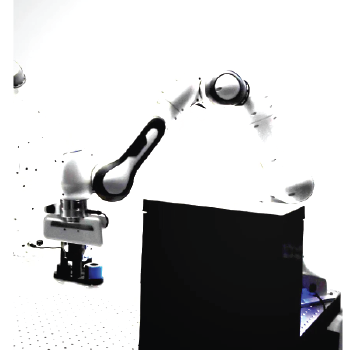}
		\includegraphics[width=0.16\textwidth]{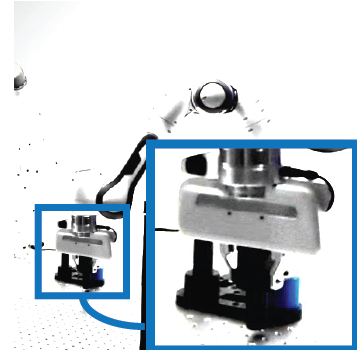}
		\label{subFig:Exp_reproGP}
	\end{subfigure}
	\caption{Snapshots of the robot reproducing the peg-in-hole task using GMR-based GP (\emph{top row}) and MOGP (\emph{bottom row}). Both methods allows the robot to circumvent the obstacle. However, the peg is successfully inserted in the hole with the GMR-based GP, while the robot fails to insert the peg with MOGP (observe that the blue hollow cylinder is behind the peg at the end of the trajectory).}  
	\label{Fig:Exp_repro}
\end{figure*}

Figure~\ref{Fig:Exp_repro} shows snapshots of the robot reproducing the peg-in-hole task using the proposed GMR-based GP (\emph{top row}) and the MOGP (\emph{bottom row}). We observe that the robot is able to circumvent the obstacle with both methods. However, the peg insertion fails when the MOGP is used, as the peg is located in front of the cylinder at the end of the trajectory. This is due to the fact that the trajectory generated by the MOGP straightly links the two zones characterized with via-points, while the GMR-based GP trajectory tends to follow the prior mean defined by GMR in between the two zones, as shown in Figure~\ref{subFig:Exp_gmm_repros}-\emph{bottom}. This behavior allows the robot to position the peg above the hole before approaching the cylinder and perform the insertion as demonstrated in the first phase of the experiment. Inversely, by using the MOGP, the robot approaches the hollow cylinder from the side, and therefore is unable to insert the peg. These different behaviors are illustrated in Figure~\ref{Fig:Exp_repro3d}, where the 3D trajectories reproduced with the different methods are represented. In order for the MOGP to successfully reproduce the insertion, a supplementary via-point could be added prior to the insertion. However, this supplementary via-point is not needed by the GMR-based GP thanks to its prior mean.

Moreover, as shown in Figure~\ref{subFig:Exp_gmm_repros}-\emph{bottom}, the uncertainty computed by the MOGP is low along the whole trajectory, resulting in a rigid behavior of the robot for the whole reproduction. In contrast, the GMR-based GP ensures a high tracking precision in the two parts of the trajectory characterized by the via-points, while the robot can be more compliant elsewhere depending on the variability encoded by the GMR, notably at the beginning of the reproduced trajectory. 
A video of the experiment and source codes are available at \href{https://sites.google.com/view/gmr-based-gp}{https://sites.google.com/view/gmr-based-gp}.

\section{Discussion}
\label{sec:Discussion}

By defining a prior mean as GMR and by encapsulating the variability of the demonstration in its uncertainty, the proposed GMR-based GP allows efficient reproductions of tasks learned by demonstration while adapting the learned trajectories towards new start-, via- or end-points.
We discuss here similarities and differences between the proposed approach and other algorithms widely used to learn trajectories. Figures accompanying the discussion are displayed in Appendix~\ref{appendix:Comparisons}.

As briefly mentioned in the previous section, adapting trajectories with GMR is difficult as conditioning on via-points results in discontinuous trajectories and re-optimizing the underlying GMM to fulfill via-points constraints is not straightforward. In contrast, the trajectories can be easily adapted to go through start-, via- or end-points by conditioning on the desired observations in the case of GP and ProMP. Trajectories can also be adapted using DMP. However, DMP does not handle variation and uncertainty information. Moreover, as DMP and ProMP encode trajectories by relying on basis functions equally spaced in time, selecting appropriate basis functions becomes difficult with high-dimensional inputs. In contrast, kernel methods and GMR, for which GMM learns the organization of basis functions, generalize well to high-dimensional inputs.
By using GP, the trajectories are modeled without considering the correlations between the output components. This problem can be alleviated by replacing GP by MOGP. 
Notice that the computational complexity of testing for the proposed GMR-based GP is importantly reduced compared to MOGP, as the set of observations used in the testing part is only composed of desired via-points, therefore resulting in a computational complexity of $\mathcal{O}(V^2 D^2)$ instead of $\mathcal{O}(N^2 D^2)$, with $D$ the output dimension and $V\ll N$.

Overall, KMP shares strong similarities with the proposed GMR-based GP. Both approaches are kernel-based and can therefore cope with high-dimensional inputs. Moreover, both make use of GMR, to retrieve a reference trajectory in the case of KMP and to define the prior mean as well as kernel parameters for GMR-based GP. Therefore, the correlations between the output components are taken into account in both models and they predict full covariances for inferred outputs. Note that both approaches can make use of other algorithms than GMR to capture the distribution of the demonstrations.

Compared to KMP which can be related to kernel regression, the framework of GMR-based GP allows the representation of more complex behaviors, notably by defining priors for the process. Our approach benefits of the properties of generative models, allowing sampling of new trajectories from prior and posterior models (as shown in Fig.~\ref{Fig:GMRbasedGPRposterior}), and is highly flexible due to the kernels $k_{\indCl}$ that can be chosen individually, resulting in different behaviors of the model in the different regions of the input space. 
Moreover, GMR-based GP provide an uncertainty information encapsulating the variability in the variance parameter of the kernel, while KMP introduces the measure of the variability of the demonstrations as the covariance matrix of the observation noise. 
As a consequence, for the example of a robot tracking via-points, KMP will adapt the distribution according to the covariance, representing the variability of the demonstrations, given initially by GMR. In contrast, our approach allows us to set via-points that the robot is required to track precisely, where the covariance tends to zero due to GP properties. This is relevant for the case in which controller gains are set as a function of the observed covariance, as we can ensure high precision due to close-to-zero prediction variances.
\vspace{-0.15cm}

\section{Conclusion}
\label{sec:Conclusion}
\vspace{-0.15cm}
This paper presented a new class of multi-output GP with non-stationary prior mean and kernel based on GMR. 
Our approach inherits of the properties of generative models and benefits of the expressiveness and versatility of GPs. Within this framework, the variability of the demonstrations is encapsulated in the prediction uncertainty of the designed GP. Correlations between the different output components are taken into account by the model.
Moreover, the method takes advantage of the prior obtained from the demonstrations for trajectory modulation, considering only via-points constraints as observed data to generate new trajectories. Our framework allows a precise tracking of via-points while the compliance of the robot can be adapted as a function of the variability of the demonstrations in other parts of the trajectories. Extensions of this work will investigate more in details the properties and limits of the proposed approach. Moreover, we plan thorough comparisons between GMR-based GP and other approaches of interest, such as KMP. Finally, the proposed approach may also be considered in an active learning framework, where new datapoints are queried in regions of high uncertainty.


\clearpage
\acknowledgments{This work was supported by the Swiss National Science Foundation (SNSF/DFG project TACT-HAND).}


\bibliography{References}  

\begin{appendices}
	\section{Comparison of GMR-based GP with ProMP and KMP}
	\label{appendix:Comparisons}
\begin{figure}[!bp]
		\centering
		\begin{subfigure}[b]{0.3\textwidth}
			\centering
			\includegraphics[width=\textwidth]{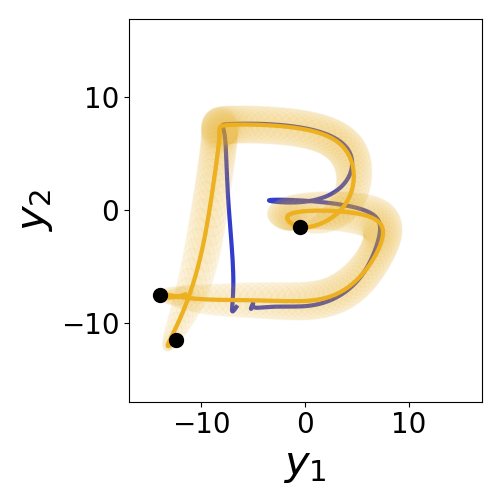}
			\caption{ProMP}
			\label{subFig:Comparison_promp}
		\end{subfigure}
		\begin{subfigure}[b]{0.3\textwidth}
			\centering
			\includegraphics[width=\textwidth]{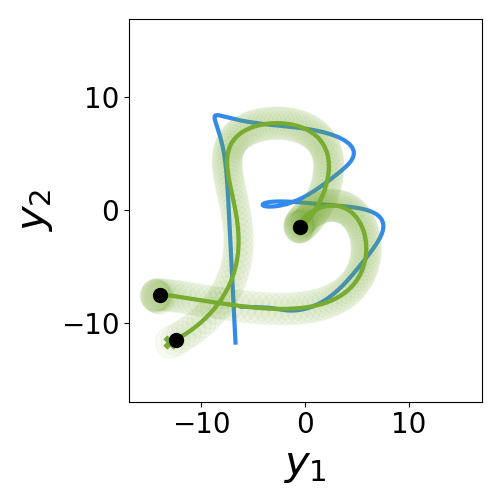}
			\caption{KMP}
			\label{subFig:Comparison_kmp}
		\end{subfigure}
		\begin{subfigure}[b]{0.3\textwidth}
			\centering
			\includegraphics[width=\textwidth]{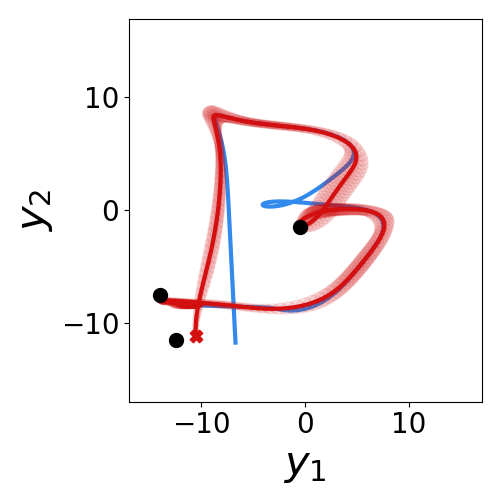}
			\caption{GMR-based GP}
			\label{subFig:Comparison_GmrBasedGPR}
		\end{subfigure}
		\caption{Comparison of the predicted trajectories generated by (\emph{a}) ProMP, (\emph{b}) KMP and (\emph{c}) GMR-based GP with three via-points. The mean is represented by a continuous line and the variance by a light tube around the estimate. Via-points are represented by black dots. The mean of the trajectories recovered from the demonstrations (without via-points) are depicted by blue lines.}  
		\label{Fig:Comparisons}
	\end{figure}

	Figure~\ref{Fig:Comparisons} shows an example of modulated trajectories recovered by ProMP, KMP and GMR-based GP. The training data, consisting of 5 demonstrations of a two-dimensional time-driven trajectory, are identical to the training data of Figures~\ref{Fig:BaseMethods} and~\ref{Fig:GMRbasedGPR}. As in Figure~\ref{Fig:GMRbasedGPRposterior}, via-points are represented by black dots. ProMP is evaluated with $20$ Gaussian basis functions, while both KMP and GMR-based GP are based on a $6$-components GMM.
	
	As expected the three methods are able to generate trajectories passing through the via-points. As discussed in Section~\ref{sec:Discussion}, ProMP encodes trajectories by relying on basis function equally spaced in time. In contrast, both KMP and GMR-based GP rely on GMR, for which GMM learns the organization of the basis functions. Note that the reference trajectory of KMP and the prior mean of GMR-based GP, depicted by a light blue line in Figures~\ref{subFig:Comparison_kmp} and~\ref{subFig:Comparison_GmrBasedGPR} are the same, as they both correspond to the mean of GMR. 
	
\begin{figure}[!bp]
		\centering			\includegraphics[width=0.25\textwidth]{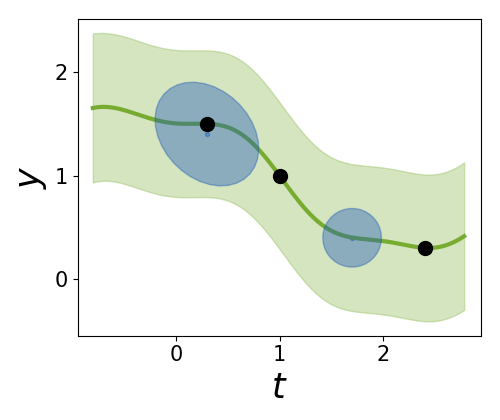}
		\caption{Example of time-driven 1D-trajectory predicted by KMP with three via-points. The reference trajectory of KMP is based on the 2-components GMM of Fig.~\ref{subFig:GmrBasedGPR_toyEx_0obs}, represented by blue ellipses. The lengthscale parameter of the Gaussian kernel is fixed as $\sigma_l=1$. The via-points are represented with black dots.}  
		\label{Fig:KMP_uncertainty}
	\end{figure}

	We observe that the global shape of the trajectory is modified by introducing via-points with KMP. In contrast, GMR-based GP tends to follow the prior trajectory in the absence of via-points. However, this change of shape allows KMP to track more precisely the first via-point compared to ProMP and GMR-based GP. Moreover, in the case of KMP, the prediction variance at the via-points depends on the variability of the GMR, while it can be set close to zero with GMR-based GP.
	This is due to the fact that, in the case of KMP, the variability of the demonstrations, encoded by GMR, is introduced as the covariance matrix of the observation noise.  In contrast, the tracking precision can be set independently with GMR-based GP, as shown by Figures~\ref{subFig:GmrBasedGPR_toyEx_3obs},~\ref{subFig:GmrBasedGPR_toyEx_noise}. 
	The aforementioned difference between KMP and GMR-based GP can also be observed by comparing Figures~\ref{Fig:GmrBasedGPR_toyEx} and~\ref{Fig:KMP_uncertainty}. Moreover, as opposite to KMP, the behavior of GMR-based GP can be modulated in different regions of the input space due to the kernel $k_\ell$ that can be chosen individually (see Figure~\ref{subFig:GmrBasedGPR_toyEx_lengthscale02}).
	
	\section{Computation Times}
	\label{appendix:CompTime}
	Table~\ref{Tab:ComputationTime} gives the computation time for one test data with the training data of the real robot experiment of Section~\ref{sec:Experiments} with a non-optimized Python code on a
	laptop with 2.7GHz CPU and 32 GB of RAM.
	
	\begin{table}[!th]
		\renewcommand*{\arraystretch}{1.1}
		\caption{Computation time of GMR, MOGP and GMR-based GP for one test data in the real robot experiment of Section~\ref{sec:Experiments}. All the time values are given in milliseconds [ms].}
		\label{Tab:ComputationTime}
		\begin{center}
			\begin{tabular}{c|c|c|}
				GMR & MOGP & GMR-based GP \\
				\hline
				$1\pm 0.1$ & $4\pm 0.6$ & $13\pm 1$ \\
				\hline
			\end{tabular}
		\end{center}
	\end{table}
	
\end{appendices}

\end{document}